\documentclass[sigconf]{acmart}
\AtBeginDocument{%
  \providecommand\BibTeX{{%
    \normalfont B\kern-0.5em{\scshape i\kern-0.25em b}\kern-0.8em\TeX}}}

\setcopyright{acmcopyright}
\copyrightyear{2018}
\acmYear{2018}
\acmDOI{XXXXXXX.XXXXXXX}

\acmConference[Conference acronym 'XX]{Make sure to enter the correct
  conference title from your rights confirmation emai}{June 03--05,
  2018}{Woodstock, NY}
\acmPrice{15.00}
\acmISBN{978-1-4503-XXXX-X/18/06}
\usepackage[ruled,vlined]{algorithm2e}

\begin{document}

\title[Improving both individual-level and system-level diversity in WeChat Feed Recommender]{Graph Exploration Matters: Improving both individual-level and system-level diversity in WeChat Feed Recommender}

\author{Shuai Yang}
\affiliation{%
  \institution{WeChat Search Department, Tencent}
  \city{Shenzhen}
  \country{China}}
\email{ariesyang@tencent.com}

\author{Lixin Zhang}
\affiliation{%
  \institution{WeChat Search Department, Tencent}
  \city{Shenzhen}
  \country{China}}
\email{lixinzhangzhang@tencent.com}

\author{Feng Xia}
\affiliation{%
  \institution{WeChat Search Department, Tencent}
  \city{Beijing}
  \country{China}}
\email{xiafengxia@tencent.com}

\author{Leyu Lin}
\affiliation{%
  \institution{WeChat Search Department, Tencent}
  \city{Beijing}
  \country{China}}
\email{leyulin@tencent.com}

\renewcommand{\shortauthors}{Trovato and Tobin, et al.}

\begin{abstract}
There are roughly three stages in real industrial recommendation systems, candidates generation (retrieval), ranking and reranking. Individual-level diversity and system-level diversity are both important for industrial recommender systems. The former focus on each single user's experience, while the latter focus on the difference among users. Graph-based retrieval strategies are inevitably hijacked by heavy users and popular items, leading to the convergence of candidates for users and the lack of system-level diversity. Meanwhile, in the reranking phase, Determinantal Point Process (DPP) is deployed to increase individual-level diverisity. Heavily relying on the semantic information of items, DPP suffers from clickbait and inaccurate attributes. Besides, most studies only focus on one of the two levels of diversity, and ignore the mutual influence among different stages in real recommender systems. We argue that individual-level diversity and system-level diversity should be viewed as an integrated problem, and we provide an efficient and deployable solution for web-scale recommenders. Generally, we propose to employ the retrieval graph information in diversity-based reranking, by which to weaken the hidden similarity of items exposed to users, and consequently gain more graph explorations to improve the system-level diveristy. Besides, we argue that users' propensity for diversity changes over time in content feed recommendation. Therefore, with the explored graph, we also propose to capture the user's real-time personalized propensity to the diversity. We implement and deploy the combined system in WeChat App's Top Stories used by hundreds of millions of users. Offline simulations and online A/B tests show our solution can effectively improve both user engagement and system revenue.
\end{abstract}







\maketitle

\begin{figure}[h]
  \centering
  \includegraphics[width=\linewidth]{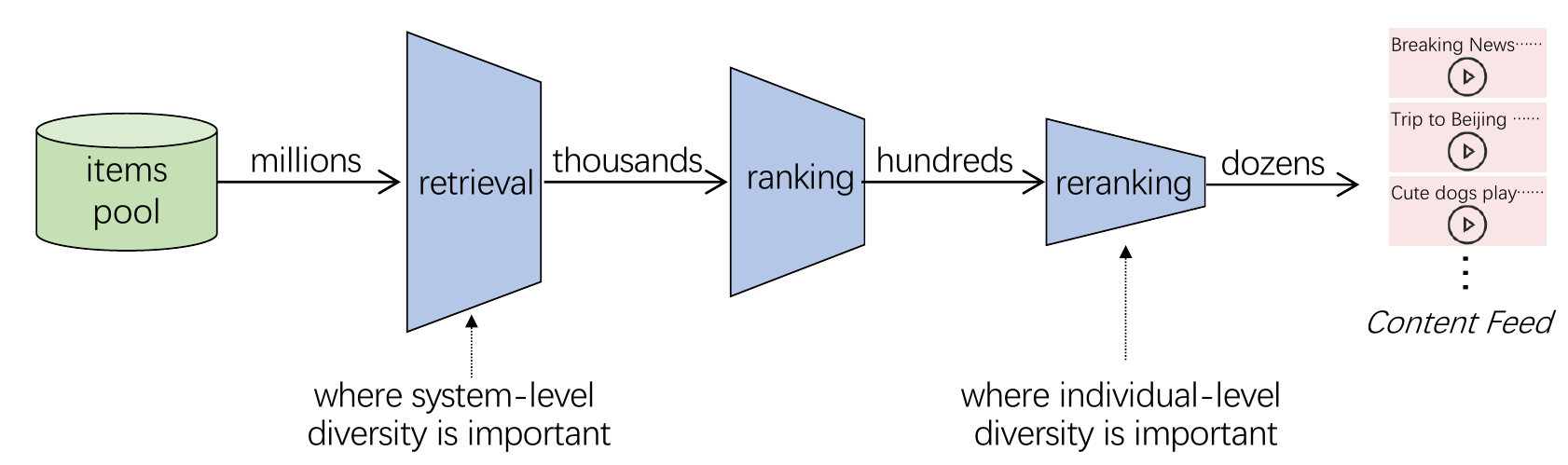}
  \caption{Recommendation stages in WeChat Top Stories.}
  \Description{overview}
  \label{pic4}
\end{figure}

\section{Introduction}
\begin{figure}[t]
  \centering
  \includegraphics[width=0.9\linewidth]{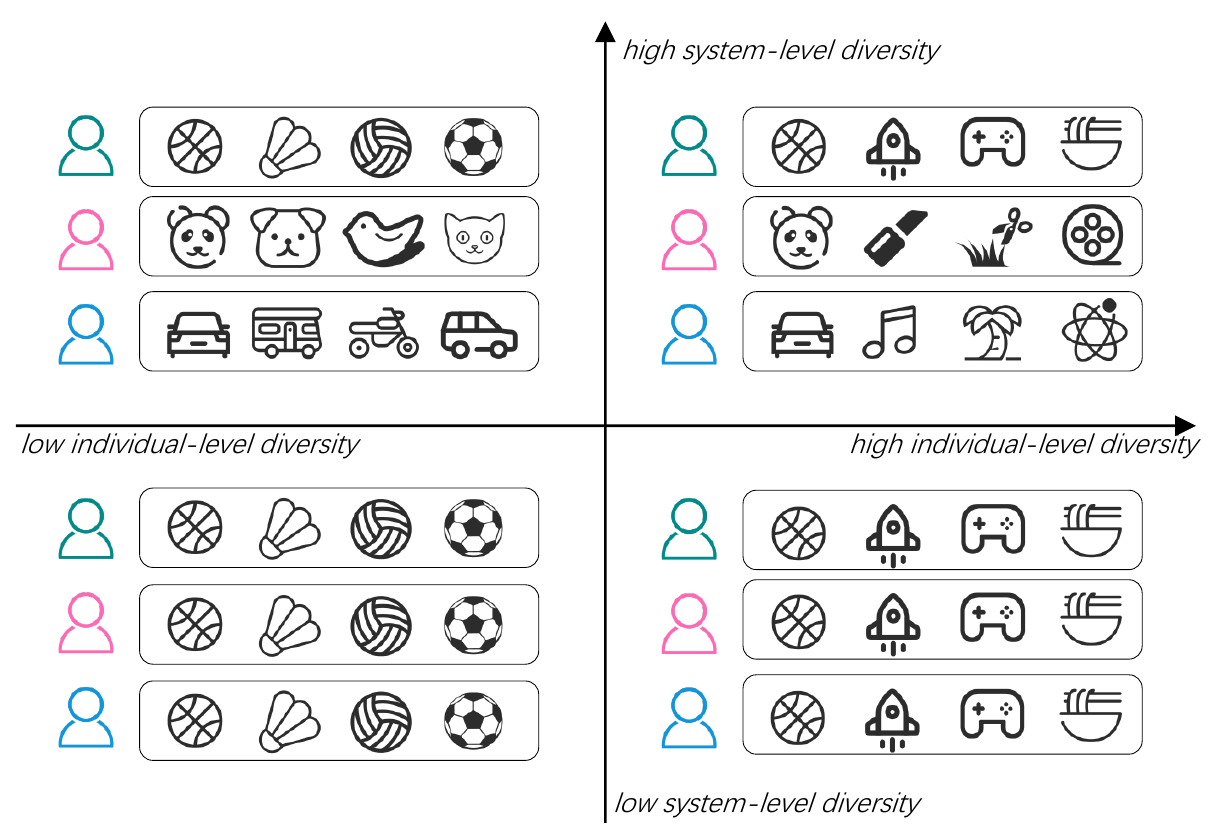}
  \caption{Both individual-level diversity and system-level diversity are important for content feed recommenders.}
  \Description{two-level of diversity}
  \label{pic3}
\end{figure}
Industrial recommender systems are widely deployed in many online services \cite{1,2,3,4,14}. WeChat, as the largest social instant messaging platform in China, also provides content feed recommendation services named Top Stories. Videos and articles are organized as a feed to be presented. Generally, most real-world recommender systems follow a standard pipeline that can by divided into three modules, namely retrieval, ranking and reranking \cite{5,6}. The main objective of the systems is to select top k items most relevant to the users. However, the output list usually contains repeated similar items, which makes users reluctant to explore the system especially in content feed scenario. Besides, those highly ranked items usually enjoy high popularity, and the diversity of each user's behavior is reduced as well. Consequently, it's less possible for new items or those with lower popularity to be exposed, and this eventually leads to the system-level convergence. Since the it influences both the user-level experience and the system-level diversity, many approaches are proposed to improve the diversity of recommender systems.

An intuitive method is to optimize the diversity in reranking layer, where ranked items are formed as the final feed exposed to users. Many algorithms, such as Maximal Marginal Relevance (MMR) \cite{10} and Determinantal Point Process (DPP) \cite{9,11,12,13}, are widely studied and deployed in real-world recommenders. DPP-based model construct a kernel matrix that contains pair-wise similarity and relevance, and shows more effectiveness and efficiency than MMR. In real-world data, different users have different diversity needs. Studies show that personalizing the diversity preference could obtain online user engagement. Recent work \cite{12} adopts behavior entropy to quantify user's propensity to diversity, and provides a fine-grained method to assign a personalized diversity factor to each user. 

However, we observe that those approaches \cite{11,12,13} usually rely on the semantic information of items, which cannot fully reveal the relevance of items, and it may fail when the category information is not accurate. The entropy-based personalization also views pair-wise distance of category as the same. While in real-world classification, "fashion" and "music" are much more close than "fashion" and "technology". Besides, by analizing the users behavior in WeChat Top Stories, we observe that users' preference to diversity changes over time. An example is in Figure 3. This may caused by individual interest shift or the distribution change of item pool, which are common cases in real-world recommenders.

Another school of thought is to improve the diversity in the retrieval stage. These methods focus on the user-wise diversity for long-term benefits. Unlike DNN-based methods \cite{14,15} focusing on training dissimilar user representations, GNN-based methods \cite{16,17,18,19,20} are devoted to learn the connections between users and items. However, without a end-to-end view of the whole system, those diversified retrieval methods could be very weak to the final diversity to the users, because the top-ranked items in the following stages are usually flooded by those with high popularity.

In this paper, we aim to improve both individual-level diversity and system-level diversity in a personalized solution. In order to overcome the limitations of existing methods, we propose a end-to-end diversification framework by making full use of graph exploration. Generally, we first adopt graph-based embedding retrieval model to match thousands of items from item pool. Then the retrieved items are sifted to dozens by the ranking stage. Based on the real-time recorded user behaviors and the graph embeddings of items, a designed function is applied to capture user's real-time propensity to the degree of diversity. In reranking stage, we view the ranked items as part of the huge graph in retrieval stage. We employ the graph embedding again to describe the similarity of ranked items. The diversified reranking problem is then converted to selecting a sub-graph with the best trade-off between relevance and diversity. With the real-time personalized diversity factor, the ranked scores and the graph-based similarity matrix, a kernel matrix can be generated for DPP to improve individual-level diversity. The selected sub-graph, is then formed as a feed for users to explore edges between diversified items. These graph exploration, in return, are merged to the huge graph to improve the system-level diversity. Since the graph-based embeddings are trained once and used multiple times in this system, our solution is very practical to be deployed in homogeneous feed recommenders.  We conduct both offline simulations and online evaluations on WeChat Topstories, which widely serves hundreds of millions of users everyday. 

The main contributions are concluded as follows:
\begin{itemize}
\item We observe and analysis that user's preference for diversity is not immutable but keep changing. We propose a function to quantify user's real-time propensity to diversity.
\item We highlight the importance of both individual-level diversity and the system-level diversity, and argue the effectiveness of end-to-end graph exploration to improve the diversity in two levels.
\item We summarize our work as a framework and implement it in a frugal manner, which has been deployed on a real-world recommendation system used by millions of users. Our work is reusable and efficient to be transplanted. 
\end{itemize}

\section{Related Work}
Throughout the evolution of recommender systems, it went through shallow stage \cite{21,22,23}, neural network stage \cite{24,25,26}, and the GNN stage \cite{16,17,18} to achieve higher accuracy. Recommendation diversity is another essential factor that impacts user experience, which is widely studied in real-world recommenders. For better comparison with our work, we summarize the use of GNN in different phases of recommendation \cite{27}, and then we elaborate on diversified solutions for recommendation. 
\subsection{GNN in Recommender Systems}
In real-world recommenders, GNN is adopted in different stages to act different roles. In retrieval phase, GNN is usually employed as the embedding trainer. The graphs can be roughly divided into user-item \cite{16,17} and item-item \cite{28,29} graphs. The graph architecture is usually constructed by analysing user's historical behaviors. These GNN-based models can capture similarity among users and items by multiple times of information propagation. Specifically, GraphSage \cite{29} has shown its capability to be deployed in web-scale recommenders \cite{28}. Combined with random walk, GraphSage conducts an inductive representation learning on large item-item graphs. It samples neighbors of the target node, aggregates and merges their embeddings with the target node to update the representations. 

Recent works also apply GNN in ranking stage, where accuracy is of first importance. GNN is employed to produce dense embeddings of the input features, and then the predictor use those embeddings to achieve ranked scores. These methods \cite{30,31} focus on the interaction among features, which is hard to capture. Compared with DNN-based methods, GNN-based ranking methods are seldom adopted in real-world feed recommendation due to the constrained accuracy and the difficulty to be deployed.

Few works adopt GNN in reranking stage, where multiple goals other than accuracy should be considered. IRGPR \cite{32} considers both item relationships and user preferences, incorporating one item relation graph to capture item relationships with another user-item scoring graph to embed personalized user intents into the propagation. However, it ignores the diversity in reranking, which is one of the most important goal in reranking stage.

\subsection{Diversified Recommendation}

\begin{figure}[t]
  \centering
  \includegraphics[width=0.9\linewidth]{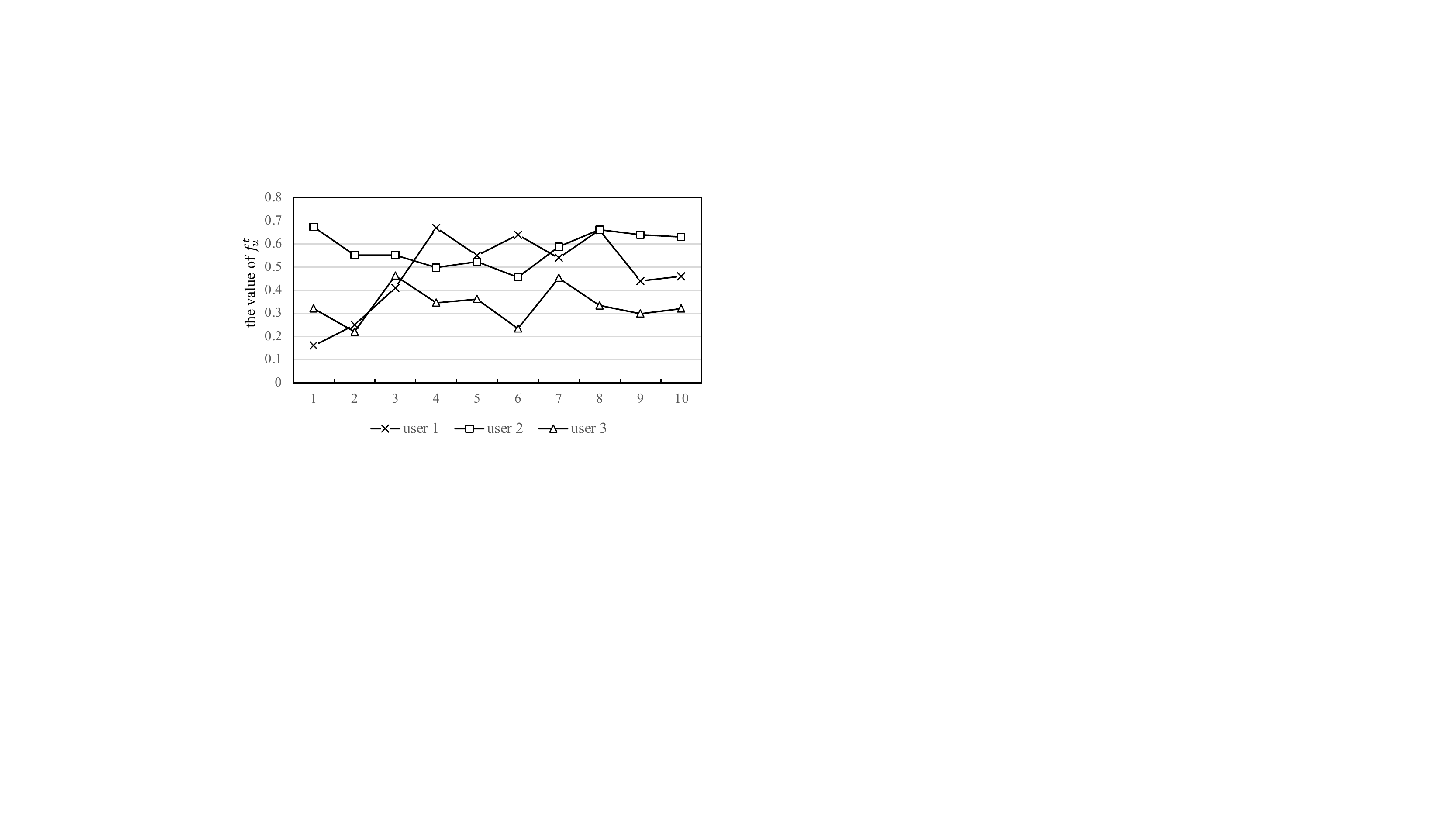}
  \caption{Three users' propensity to the degree of diversity changing in 10 days. Values are $f_{u}^t$ explained in section 3.2. A larger value means user prefers more similar items.}
  \Description{user's preference for diversity changes over time.}
  \label{pic5}
\end{figure}

Recommendation diversity should be measured in two typical dimensions: individual-level diversity and system-level diversity \cite{33,34,27}. Individual-level diversity measures the dissimilarity of the recommended items for each user, whereas system-level diversity measures the dissimilarity of recommendation results of different users. Both levels of diversity are important and should be considered as an integrated problem.

In order to improve individual-level diversity, diversified reranking has been widely studied. Maximal Marginal Relevance (MMR) \cite{10} method considers both the relevance and the pair-wise similarity of ranked items by point-wise rerank. DPP is an elegant probabilistic model with the ability to express negative interactions. DPP-based models \cite{7,9,11,12,13} construct a kernel matrix that contains pair-wise similarity and relevance, and shows more effectiveness and efficiency than MMR. Fast greedy maximum a posteriori inference for DPP (Fast-DPP) \cite{11} makes it easy to be deployed in web-scale recommender systems. In real-world data, different users have different diversity needs \cite{35,36}. Studies \cite{37,38} show that personalizing the diversity preference could obtain online user engagement. Based on DPP, \cite{12} adopts behavior entropy to quantify user's propensity to diversity, and provides a fine-grained method to assign a personalized diversity factor to each user. It relies heavily on the pre-defined semantic information, which cannot reveal the interaction similarity. In our work, we observe that users' preference to diversity changes over time. This may caused by individual interest shifting or the distribution change of item pool, which are common cases in real-world recommenders. Our approach provides a function to capture both the user interests shifting and the item pool change, and simply adopts Fast-DPP in reranking stage to improve individual-level diversity.

However, as the last strategy before exposure, diversified reranking methods focus on improving item-wise diversity showed to users but overlook the system-level diversity, which is much more important for long-term gain. In fact, retrieval should take more responsibility for diversity, since it emphasizes the coverage of user-interested items rather than their specific item ranks. Luckily, some works have been studied to improve the retrieval diversity. For example, DNN-based method MIND \cite{14} adopts a multi-interest extractor layer with a variant dynamic routing in neural network to extract user’s diverse interests. GNN-based methods are also modified to capture long-tail items in graph. GraphDR \cite{19} builds a huge heterogeneous preference network to record different types of user preferences, and conduct a fieldlevel heterogeneous graph attention network for node aggregation. Both DGCN \cite{39} and DGRec \cite{20} focus on rebalancing the categories of retrieved items, which also relies heavily on the pre-defined semantic information. As mentioned previously, these graph-based models are hard to be deployed in ranking or reranking stage, so in real-world web-scale recommeders they can just serve in retrieval stage.

These methods focuses on either individual-level diversity or system-level diversity separately. Our approach views the individual-level diversity and the system-level diversity as an integrated problem, and attempts to solve it in a simple yet effective solution. We make full use of the item-item graph embedding in retrieval stage to capture use's real-time preference to diversity, reveal the interactive similarity of ranked items in reranking stage. A sub-graph with the best trade-off between relevance and diversity is generated by real-time personalized DPP, and the path is explored by real-world users. Those graph exploration are then merged in the huge item-item graph to increase the system-level diversity.

\begin{figure}[t]
  \centering
  \includegraphics[width=\linewidth]{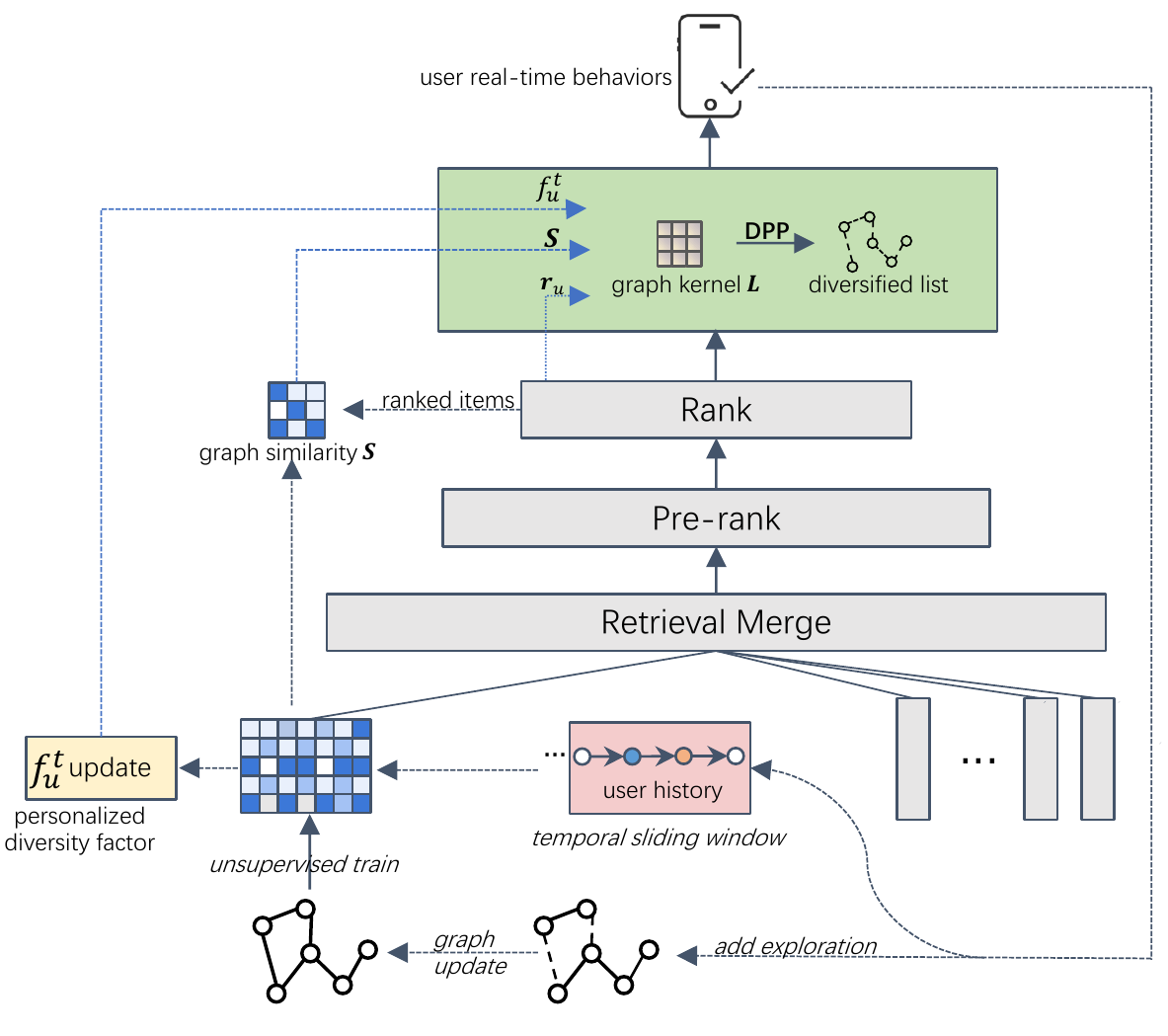}
  \caption{The overview of our solution for two levels of diversity. We integrate the retrieval stage and the reranking stage by using the item-item graph.}
  \Description{A woman and a girl in white dresses sit in an open car.}
  \label{pic1}
\end{figure}

\section{METHOD}
\subsection{GNN-based Retrieval}
GraphSage \cite{28} allows embeddings to be efficiently generated for unseen nodes by leveraging node feature information. It has been successfully applied to web-scale recommender systems to compute embedddings for billions nodes. It samples neighbors of one node, aggregates neighbourhood embeddings to update node representations.
\begin{equation}
\begin{aligned}
\mathbf{h}_{\mathcal{N}_i}^l & =\operatorname{AGGREGATE}_l\left(\left\{\mathbf{h}_j^l, \forall j \in \mathcal{N}_i\right\}\right), \\
\mathbf{h}_i^{l+1} & =\delta\left(\mathbf{W}^l\left[\mathbf{h}_i^l \| \mathbf{h}_{\mathcal{N}_i}^l\right]\right)
\end{aligned}
\end{equation}
where The embedding of graph node $i$ in the $l$-th propagation layer, $\mathcal{N}_i$ is the neighborhood set of node $i$, $\delta$ is the activation function, $\mathbf{W}$ is trainable matrix ,$\|$ is the concatenation operation.
In our work, we view the items as nodes in the graph, and the edge is connected if two items are positively interacted in one user's behavior session. Item attributes are used as node features. Random walk is adopted to sample neighbours of each node. Then we use the unsupervised GraphSage to generate the embeddings of all the items. Based on user's historical positive interacted items, we can easily match the potential items by computing the cosine similarity of item embeddings.

\subsection{Real-time Personalization}
With the observation of the temporal change of user's preference to individual-level diversity, we propose a scheme to capture user's real-time propensity to the degree of diversity, which is personalized as well. 

First, a simple yet effective temporal sliding window is applied to filter user's recent positively interacted items $\mathbf{I}_u^t$. With the help of the huge graph in retrieval stage, the similarity matrix $\mathbf{S}_{\mathbf{I}_u^t}$ of these selected items from user $u$ can be computed with normalized embeddings $\mathbf{e}_u^t$, and we use $s_{\mathbf{u}}^t$ as the mean item similarity of user in time $t$:
\begin{equation}
\begin{aligned}
    \mathbf{S}_{\mathbf{I}_u^t} = \left( \mathbf{e}_u^t * \mathbf{e}_u^t\top\right), \\
    s_{\mathbf{u}}^t = \operatorname{mean}_{i, j \in \mathbf{I}_u^t, i \neq j}\left(1-\mathbf{S}_{{\mathbf{I}_u^t}_{i j}}\right)
\end{aligned}
\end{equation}

Then $s_{\mathbf{u}}^t$ is updated in a vector $\mathbf{s}^t$, which records all users' real-time item similarity. We adopt z-score to standardize all users' $s_{\mathbf{u}}^t$, then sigmoid function is used to return a value $f_{\mathbf{u}}^t$ in the range 0 to 1. 
\begin{equation}
 f_{u}^t = \frac{1}
             {1 + \exp \left( \frac{-(s_u^t - \bar{\mathbf{s}^t})}
                                   {\sigma_{\mathbf{s}^t}}
                       \right)}   .
\end{equation}
According to \cite{40}, user characteristics are correlated with their historical behaviors in recommenders. We can simply make the assumption that the conclusion that user $u$ with a large $f_{u}^t$ at time $t$ tends to view more similar items. Thus this value can be viewed as user's real-time propensity to the degree of diversity. It will be more explainable when combined with fast DPP in the next subsection.

\begin{algorithm}[t]
	\caption{Reranking Stage.}
	\label{alg:algorithm1}
	\KwIn{: Ranked items set: $Z$; Real-time personalized preference for diversity of user $u$: $f_{u}^t$; Graph-based similarity matrix of ranked items: $\mathbf{S}$; Ranked scores of items: $\mathbf{r}$; Output size $k$.}
	\KwOut{Reranked top-k items set $Y$.}  
	\BlankLine
        Compute intermediate scores $\mathbf{r}^{*}$ by equation (7).
        
	Compute kernel matrix $\mathbf{L}$ by equation (6) using $\mathbf{r}^{*}$ and $\mathbf{S}$ .
 
        Initialize $Y = \{\}$.
        
	\While{\textnormal{$|Z| > 0$ and $|Y| < k$}}{
 
		Select item $j$ by equation (8).
  
	    $Y \cup\{j\}$.
	}
 
	Return: $Y$.
 
\end{algorithm}

\subsection{Reranking}

It's natural to adopt fast-DPP [11] in large sale recommendation systems for efficiency. Positive semi-definite kernel matrix $\mathbf{L}$ can be generated by using relevance scores $\mathbf{r}$ (usually the ranked score) and normalized item embeddings $\mathbf{f}$ by
\begin{equation}
\begin{aligned}
    \mathbf{L}_{ij} = r_{i}r_{j}<\mathbf{f}_i,\mathbf{f}_j>, \\ 
    \mathbf{L}=Diag(\mathbf{r}) * \mathbf{S} * Diag(\mathbf{r})   
\end{aligned}
\end{equation}
where $\mathbf{S}$ is the similarity matrix of the ranked items. The log-probability of selected item subset $Y$ can be written as
\begin{equation}
    log\mathcal{P}(Y) \propto \theta * \sum \limits_{i \in Y} r_i + (1-\theta) * log(det(\mathbf{S}_{Y})), \theta \in (0,1)
\end{equation}
where $det$ is the determinant of matrix, $\theta$ is the trade-off parameter between relevance and diversity. A larger $\theta$ means more relevance of results. When $\theta = 0.5$, users obtain a equal importance between relevance and diversity. The corresponding kernel with trade-off can be written as:
\begin{equation}
    \mathbf{L}^{'} =Diag(exp(\alpha\mathbf{r})) * \mathbf{S} * Diag(exp(\alpha\mathbf{r})), \alpha=\theta/2(1-\theta).
\end{equation}

Study in \cite{40} has shown that the historical behavior can reflect user's preference for diveristy. $f_u^t$ mentioned in section 3.2 forces the user similarity in a distribution ranging in 0 to 1, which follows the fact that less users prefer either extremely more diversity or relevance . Thus we use $f_u^t$ as the real-time trade-off parameter $\theta$ between relevance and diveristy. In practice, the trade-off can be considered by rewriting each ranked score as:
\begin{equation}
    r^{*}_i=exp^{\frac{f_{u}^t}{2(1-f_{u}^t)}*r_i}, i \in Z.
\end{equation}

\begin{figure*}[t]
  \centering
  \includegraphics[width=\textwidth]{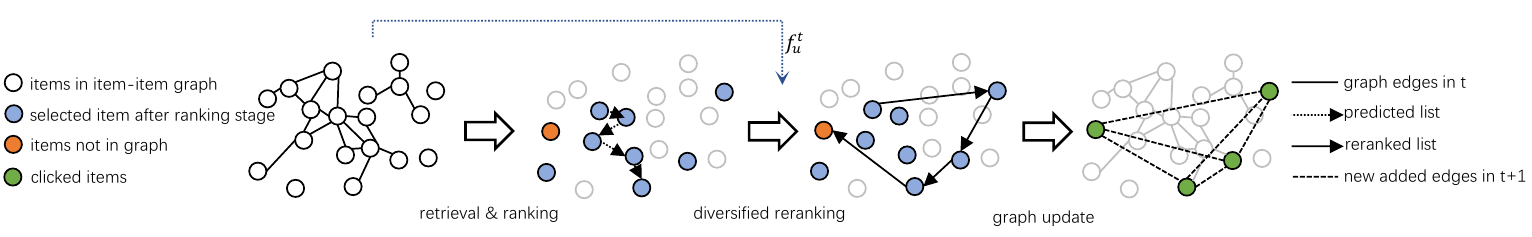}
  \caption{An illustration of graph exploration in proposed framework. An edge between two nodes means they are both positively interacted by the same user in one recommendation session. Shorter distance between items means higher similarity.}
  \Description{A woman and a girl in white dresses sit in an open car.}
  \label{pic2}
\end{figure*}

Instead of using category-like semantic embedding of items as the input of similarity computation, we again employ the embeddings from the huge item-item graph to generate the similarity matrix $\mathbf{S}$ of ranked items. The embeddings contain both item features information and their interacted information. The former focuses on individual-level diversity, whereas the latter is more important for system-level diversity. 

Given the similarity matrix $\mathbf{S}$, the ranked scores $\mathbf{r}$ and the real-time personalized diversity preference $f_{u}^t$, the kernel matrix $\mathbf{L}$ can be generated. Fast-DPP converts problem of selecting subset $Y$ from whole set $Z$ to greedily selecting item $j$ with the most promotion to the determinants of the updated submatrix iteratively, since $\mathcal{P}(Y) \propto det(\mathbf{L}_{Y})$: 
\begin{equation}
j=\arg \max _{i \in Z \backslash Y} \log \operatorname{det}\left(\mathbf{L}_{Y \cup\{i\}}\right)-\log \operatorname{det}\left(\mathbf{L}_{Y}\right)
\end{equation}
We formalize the deployed reranking stage in Algorithm 1.

\subsection{Graph Exploration}
With the components deconstructed above, we now explain how our framework improve the individual-level diversity and system-level diversity in an integrated manner. An example of top-5 items recommendation task is illustrated in Figure 5.

\textbf{Retrieval and ranking}. 
The graph embeddings are online trained by using real-time feedback from users. The graph embeddings are first used as a retrieval strategy. After ranking, hundreds of selected items can be viewed as nodes in the large graph. The ranked scores predict top-k items as the most relevant items and usually very similar to user's strong interest. Thus a higher coverage of items in retrieval stage is needed.

\textbf{Real-time personalized trade-off between relevance and diversity}. 
As expounded in section 3.2, combined with user's history, $f_{u}^t$ is also generated by the graph embeddings. This parameter contains user's diversity information based on the graph. It determines the degree of graph exploration.

\textbf{Explore unseen items}. The ranked items are not necessarily in the graph, because in real-world recommenders there are dozens of retrieval strategies other than graph-based retrieval. Luckily GraphSage can predict the unseen items' embedding by using the node features. In practice, the feature space is usually the same as the existed graph nodes. Then the embedding can be predicted without edges connected. Thus the unseen items can also be viewed as part of the huge graph.

\textbf{Explore unconnected edges}. The graph is again used to generate the similarity matrix of ranked items, which contains node features similarity and interaction-level similarity. With the help of fast-DPP, top-k items are re-selected. Since node features contain semantic and category-like information, the diversified reranked items involve more semantic diversity, which improves the individual-level diversity. Also, since the graph embedding contain interaction-level similarity, items, usually together positively interacted by the same user, are dispersed after DPP. The latter aims to improve the system-level diversity by attempting to add new edges in the graph.

\textbf{Graph update}. The reranked items are then exposed to user as a feed. Items with positive interaction in the same session are defined as correlated pairs, and the edges are then added in the huge item-item graph. The recommender now completes one round of graph exploration, and the graph is continuously iterated after each recommendation lifecycle. In the next round, with the added edges and nodes, the graph-based retrieval matches more diversified items, which indicates a higher coverage of items and exactly improves the system-level diversity Graph exploration can improve both levels of diversity by leveraging the "learning effect" mentioned in \cite{41}.

\subsection{Real-world Deployment}
The graph-based retrieval model is implemented in C++ as an incremental learning version to be easily updated. User's behavior history are also updated in a real-time manner. $s_{\mathbf{u}}^t$, which records all users' real-time item similarity based on behavior history applied with a temporal sliding window, is also updated in a real-time way. The size of the sliding window can be defined as the space limit or the time limit. We find the latter works better, because hot events often intensively change the data distribution in content feed recommendation task. These real-time updated parameters, together with the ranking models, are stored in online dict severs to ensure the accuracy and diversity of the recommender system. The whole framework reuse the graph embedding 3 times respectively in retrieval stage, $f_{u}^t$ generation, and the similarity matrix generation in reranking stage, which doesn't rely on extra pre-trained embeddings or semantic information. This is a saving solution, and it prevents the reranking DPP suffering from clickbait or inaccurate categories, which is the common case in content feed recommendation.

\section{Experimental results}

Two stages of experiment are conducted in WeChat Top Stories videos recommendation. We first conduct a 7-days online experiment to compare methods with each other, and then the best method is launched for a long-term observation. This is a frequently-used method in industrial recommenders. In this section, we first prove the effectiveness of our work compared with others, and then show the long-term gain since launched.

\begin{table}[b]
\caption{Experimental Results in One Week}
\begin{tabular}{@{}llllll@{}}
\toprule
method   & IDL             & RD              & VV              & Staytime        & AD              \\ \midrule
Base     & -               & -               & -               & -               & -               \\
fast-DPP & +0.341          & *               & +0.912          & *               & +0.316          \\
p-DPP    & +0.367          & *               & +1.117          & +0.391          & +0.684          \\
ours     & \textbf{+2.003} & \textbf{+4.678} & \textbf{+2.397} & \textbf{+0.725} & \textbf{+1.143} \\ \bottomrule
\end{tabular}
\label{tab:result1}
\end{table}

\begin{figure*}[h]
  \centering
  \includegraphics[width=0.99\textwidth]{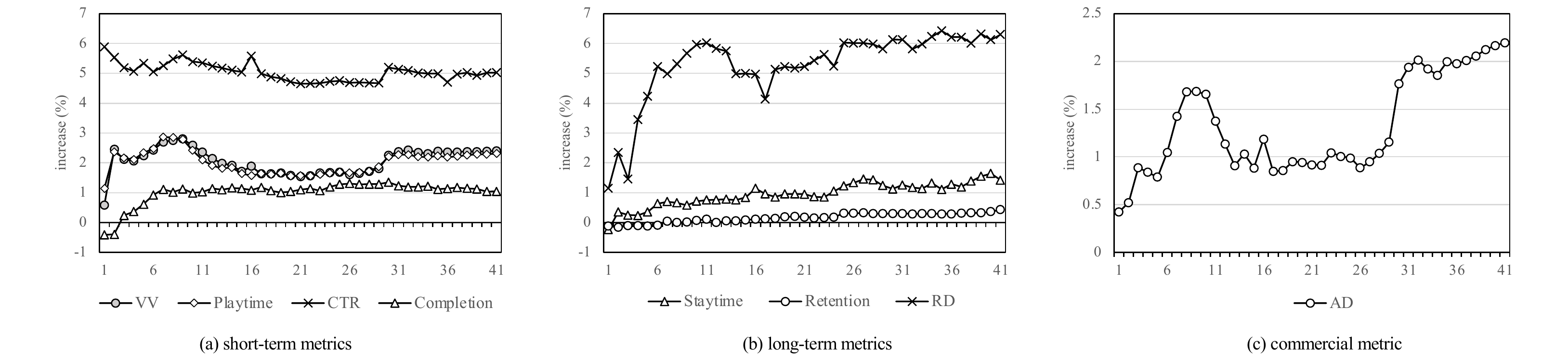}
  \caption{An illustration of graph exploration in proposed framework. An edge between two nodes means they are both positively interacted by the same user in one recommendation session. Shorter distance between items means higher similarity.}
  \Description{A woman and a girl in white dresses sit in an open car.}
  \label{pic6}
\end{figure*}

\textbf{Baselines}.
Three baselines are compared in the 7 days online experiment. We refer the existed rule-based diversified reranking as "Base", which limits the number of items with the same categort or tags. Simply adopting fast-DPP \cite{11} based on the category-like similarity is referred as "fast-DPP". The solution proposed in \cite{12} is p-DPP, which considers the personal difference of propensity to diversity. 

\textbf{Metrics}.
We adopt intra-list average distance (ILAD) \cite{11} to measure the individual-level diversity, which is the average of all users' mean distance of interacted items:
\begin{equation}
\operatorname{ILAD}=\operatorname{mean}_{u} \operatorname{mean}_{i, j \in R_u, i \neq j}\left(1-\mathbf{S}_{i j}\right)
\end{equation}
where $R_u$ denotes the item lists exposed to user $u$. 
And we propose to measure the system-level diversity by computing the average distance of retrieval results (RD), which is similar to ILAD in computation. In our experiment, IDL is computed based on the item-item graph embeddings. We adopt a series of important metrics in feed recommendation. To measure short-term performance, per capita video views (VV), per capita playtime (Playtime), per capita click through rate (CTR), and play completion rate (Completion) are adopted. For long-term evaluation, 
per capita staytime (Staytime) and weekly retention rate (Retention) are adopted. Besides, we also evaluate the commercial revenue by 
per capita advertisements exposure (AD). The p-value is set as 0.05. Online A/B experiments are conducted on buckets of users, and the results are daily collected.

\textbf{Compared with baselines}.
The results of a 7-days online experiment are showed in Table ~\ref{tab:result1}. A "*" means null hypothesis and can be viewed the same as the baseline. Compared with the rule-based diversification, DPP-based methods can significantly improve the individual-level diversity. Meanwhile, the diversified items can improve users' consumption in feed, which directly demonstrates the effectiveness of DPP-based reranking. Compared with other DPP-based methods, our work improves the retrieval diversity, which reveals the effectiveness of graph exploration. This verifies that our framework can simultaneously improve both individual-level diversity and system-level diversity without reducing the accuracy. Our method gains the most online increase, and is then launched for long-term observation.

\textbf{Long-term experiment}.
As is shown in Figure ~\ref{pic6}, our method obtains consistent increase in several aspects compared with the "Base" method. For short-term metrics, our method can significantly improve the click-through rate, which reveals that diversified top-k items are more likely to be positively interacted than those similar items. The observed growth in VV, Playtime, and Completion also confirms this conclusion. We can observe the learning effect in long-term gains. Users tend to spend more time in our feed, and the retention rate also increase in a growing manner. The RD is also significantly improved especially in the first few days, because the graph in trained in an incremental way. With more graph exploration, the retrieval diversity is continuously growing and it consequently affects the diversity in reranking stage. This verifies that graph exploration can obtain a long-term gain in system-level diversity. Besides, the commercial metric AD also increases, because advertisements are regularly inserted in our feed. This proves that our work can bring business growth.

Both of the experiments demonstrate that individual-level diversity and system-level diversity can affect mutually when it comes to industrial feed recommenders. Thus these two levels of diversity should be viewed as an integrated problem. Graph exploration can efficiently and significantly improve both of them without losing the online performance. Besides, compared with methods training and loading another semantic embedding table, the reuse of graph embedding saves storage and computing resources.

\section{Conclusions}
Researchers focus on either individual-level diversity or system-level diversity in one local stage of the recommendation systems, which ignores the mutual influence with other stages. This makes it hard to achieve improvement in standard three-stage industrial recommenders. In this paper, we treat proving two levels of diversity as an integrated task from a systematic and dynamic perspective. We discuss the importance of graph exploration in both retrieval and reranking stage. The proposed framework reuses the graph in different phases of recommender in an efficient way. With the natural iteration of the system, our work obtains consistent improvement of short-term, long-term, and commercial benefits in WeChat Top Stories.  Besides, we address that users' propensity to the degree of diversity changes over time, and a real-time personalization strategy is adopted to achieve a better trade-off between relevance and diversity. Our work focuses on the practical issues of industrial recommenders, and provide an efficient and deployable solution.

\bibliographystyle{ACM-Reference-Format}
\bibliography{yang}


\end{document}